\documentclass{article}

     \PassOptionsToPackage{numbers,sort,compress}{natbib}

\usepackage[preprint]{neurips_2021}

\usepackage[utf8]{inputenc} 
\usepackage[T1]{fontenc}    
\usepackage{url}            
\usepackage{booktabs}       
\usepackage{amsfonts}       
\usepackage{nicefrac}       
\usepackage{microtype}      
\usepackage{xcolor}         
\usepackage{times}
\usepackage{epsfig}
\usepackage{graphicx}
\usepackage{amsmath}
\usepackage{bbm}
\usepackage{amssymb}
\usepackage{graphicx,times,amsmath}
\usepackage{cases}
\usepackage{multirow}
\usepackage[ruled,vlined]{algorithm2e}
\usepackage{amsfonts}
\usepackage{graphicx}
\usepackage{amssymb}
\usepackage{amsfonts}
\usepackage{amsthm}
\usepackage{subfig}
\usepackage{epstopdf}
\usepackage{makecell}
\usepackage{array}
\usepackage{xspace}
\usepackage{xparse}
\usepackage{mathtools}
\usepackage{makecell}
\usepackage[symbol]{footmisc}
\usepackage{rotating}
\usepackage{comment}
\usepackage{booktabs}
\usepackage{wrapfig}

\makeatletter
\newcommand{\nosemic}{\renewcommand{\@endalgocfline}{\relax}}
\newcommand{\dosemic}{\renewcommand{\@endalgocfline}{\algocf@endline}}
\newcommand{\midsepremove}{\aboverulesep = 0mm \belowrulesep = 0mm}

\makeatother

\newcolumntype{?}{!{\vrule width 1pt}}

\DeclareMathOperator*{\argmin}{arg\,min}
\newcommand\doubleplus{+\kern-1.3ex+\kern0.8ex}

\usepackage[colorlinks=true,linkcolor = red, breaklinks=true,letterpaper=true,bookmarks=false]{hyperref}

\makeatletter
\DeclareRobustCommand\onedot{\futurelet\@let@token\@onedot}
\def\@onedot{\ifx\@let@token.\else.\null\fi\xspace}
\def\eg{\emph{e.g}\onedot} 
\def\ie{\emph{i.e}\onedot} 
 
\def\etc{\emph{etc}\onedot}

\makeatother

\title{TransCamP: Graph Transformer for 6-DoF Camera Pose Estimation}
\author{%
Xinyi Li\\
Dept. of Computer \& Information Sciences\\
Temple University\\
Philadelphia, PA 19122\\
{\tt\small xinyi.Li@temple.edu}\\

\And
Haibin Ling\\
Dept. of Computer Science\\
Stony Brook University\\
Stony Brook, NY 11794\\
{\tt\small hling@cs.stonybrook.edu}
}
\begin{document}

\maketitle
\begin{abstract}
Camera pose estimation or camera relocalization is the centerpiece in numerous computer vision tasks such as visual odometry, structure from motion (SfM) and SLAM. 
In this paper we propose a neural network approach with a graph transformer backbone, namely TransCamP, to address the camera relocalization problem.
In contrast with prior work where the pose regression is mainly guided by photometric consistency, TransCamP effectively fuses the image features, camera pose information and inter-frame relative camera motions into encoded graph attributes and is trained towards the graph consistency and accuracy instead, yielding significantly higher computational efficiency. 
By leveraging graph transformer layers with edge features and enabling tensorized adjacency matrix, TransCamP dynamically captures the global attention and thus endows the pose graph with evolving structures to achieve improved robustness and accuracy. 
In addition, optional temporal transformer layers actively enhance the spatiotemporal inter-frame relation for sequential inputs. 
Evaluation of the proposed network on various public benchmarks demonstrates that TransCamP outperforms state-of-the-art approaches.
\end{abstract}

\section{Introduction}
\label{sec:intro}
The past decade has witnessed surging research interest in developing camera relocalization methods, benefiting various computer vision applications including robot navigation, autonomous driving and AR/VR technologies. Conventional approaches solving the camera pose estimation problem involve extensive implementations of bundle adjustment~\cite{triggs1999bundle}, which is the iterative process of joint optimization of the 3D scene points and the 6-DoF camera pose parameters, aided by numerical solvers. The formulation yields a non-linear high-dimensional system and is thus computationally challenging to solve.

With the prevalence of deep neural networks, many recent studies have steered research attentions towards leveraging deep learning techniques to re-formulate the camera pose estimation problem as a pose regression network, \ie, the network is trained with training images and the ground-truth camera poses such that it can learn to regress the camera pose(s) given single or multiple images. Among these research, PoseNet~\cite{kendall2015posenet} pioneers in incorporating neural networks into camera pose regression frameworks, where the CNN-based network is trained to directly estimate the camera pose from individual images without explicit feature processing. 
Later work includes VidLoc~\cite{clark2017vidloc} in which a CNN-RNN joint model is presented to address the temporal consistency of the sequential images. Recently GNNs have been exploited in a camera pose regression framework~\cite{xue2020learning}, where the message passing scheme enables obtaining the inter-frame dependency.

Lately, the development of Transformers~\cite{vaswani2017attention} has empowered massively successful applications in natural language processing (NLP), computer vision and many other fields. Specifically, the adoption of the self-attention mechanisms enables Transformer to effectively capture the global spatiotemporal consistency of sequential information. 
Additionally, while graph-based networks such as GNNs have been widely proven to be efficient in modeling arbitrarily structured inputs, it is generally computationally challenging to have the networks update the graph structure dynamically, limiting its performance on downstream tasks where high amounts of noise or missing information are present.

Inspired by the aforementioned observations, in this work we propose a neural network fused with a graph transformer backbone, namely \textit{TransCamP}, to tackle the camera pose estimation problem. In TransCamP, the view graph is constructed by a novel graph embedding mechanism, where the nodes are encoded with image features and 6-DoF absolute camera pose of the frame while the edge attributes consist of the relative inter-frame relative camera motions. Moreover, our proposed network introduces a tensorized adjacency matrix that stores the correlation on both the feature level and the frame level. In particular, the feature correspondences between the frames are encoded into the element in the adjacency matrix, and the element value is based on the normalized feature correspondence score and thus falls in the range of $[0, 1]$. The adjacency matrix is updated through the graph transformer layers to reflect the evolving graph structure, \eg, redundant/noisy edge pruning,  newly-added edges according to high correlations between a new image and some previous image, \etc. TransCamP is trained end-to-end, guided by the loss function that integrates the graph consistency. Additionally, the temporal transformer layers are utilized to obtain the temporal graph attention for consecutive images.

\begin{figure}[t!]
    \centering
    \includegraphics[width=\linewidth]{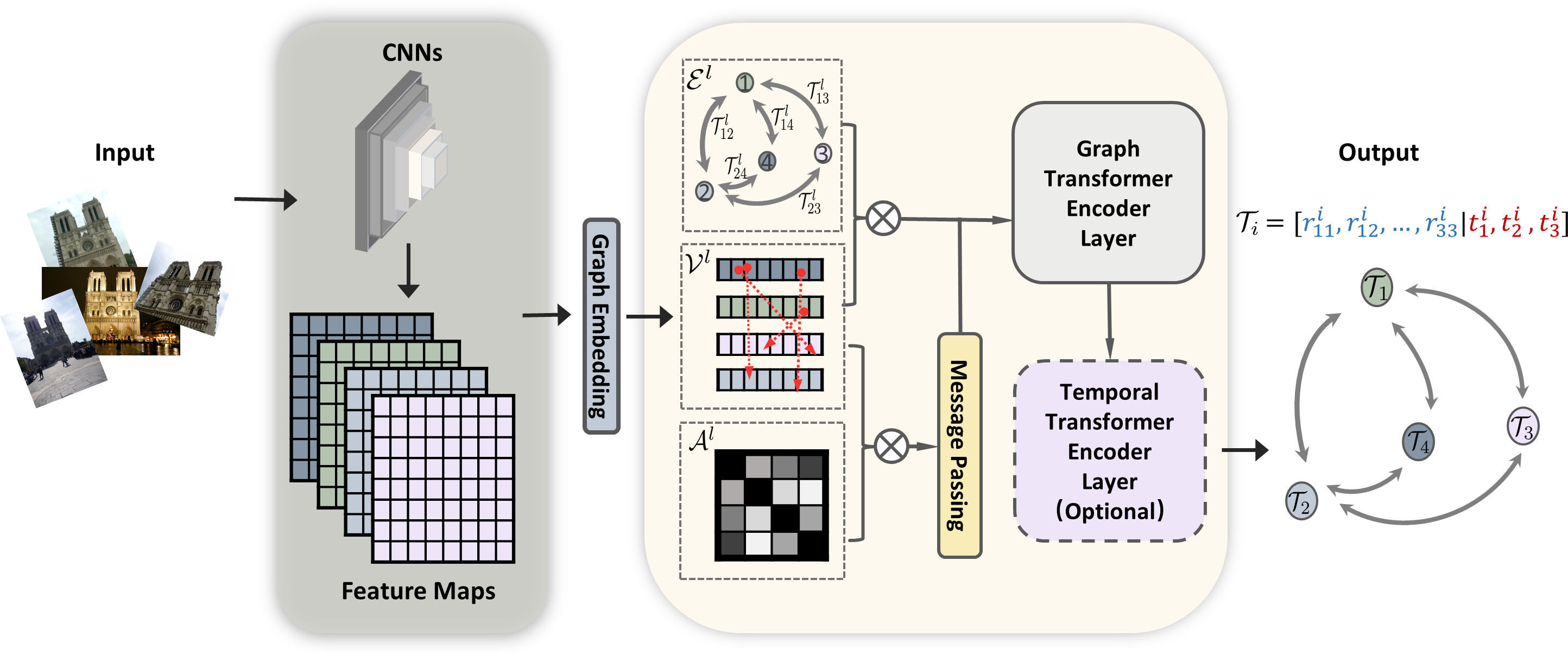}
    \caption{Overview of the proposed TransCamP architecture for real-time camera relocalization. The network takes images as input and then models the corresponding camera poses, image features and the pair-wise relative camera motions into a graph $\mathcal{G}(\mathcal{V}, \mathcal{E})$. Then, the adjacency tensor $\mathcal{A}$ and nodes are fed into the message passing layers, before passing through the graph transformer encoder layers. For consecutive image sequences, the graph will be passed through additional temporal transformer encoder layers. The global camera poses are embedded into the node information in the final output.}
    \label{fig:pipeline}
\end{figure}

The architecture overview of TransCamP is given in Fig.~\ref{fig:pipeline}. The design of the proposed network is favorable for camera relocalization tasks in three aspects. First, it is efficient to exploit the intra- and inter-frame structure information and correlation with the utilization of graphs; Second, the self-attention mechanism can effectively capture the spatiotemporal consistency in arbitrarily long-term periods, achieving high global pose accuracy; Third, with the adjacency matrix being dynamically updated, the network can quickly adjust according to the changing graph structure, further reducing the negative effects caused by erroneous feature matching. 

To the best of our knowledge, our proposed network is the first to exploit Transformer for the camera relocalization task. We will release the code upon the acceptance of the paper. 

\section{Related Work}
\label{sec:relwork}
\textbf{Graph Transformers.} By virtue of its powerful yet agile data representation, GNNs~\cite{scarselli2008graph, velivckovic2017graph, kipf2016semi} have achieved exceptional performances on numerous computer vision tasks. Despite {their} successes, straightforward adoptions of GNNs in modeling camera relocalization task is not applicable due to GNN's vulnerability against noisy graphs~\cite{gidaris2019generating, xu2018powerful, zhao2019pairnorm, rong2019dropedge}. In~\cite{devlin2018bert}, Graph-BERT enables pre-training on the original graphs and adopts a subgraph batching scheme for parallelized learning. However, Graph-BERT assumes that the subgraphs are linkless, \ie, there exists no edge connecting subgraphs, thus not suitable for tasks where global connectivity is important.
Recently with the success of Transformers~\cite{vaswani2017attention}, several work~\cite{yun2019graph, dwivedi2020generalization, wang2019heterogeneous} have attempted to develop graph transformers which can leverage the powerful message passing scheme on graphs while utilizing the multi-head self attention mechanism in Transformers. Among which the approach proposed in~\cite{wang2019heterogeneous} is capable of transforming the heterogeneous graphs into homogeneous graphs such that the transformer can be exploited.
GTNs proposed in~\cite{yun2019graph} also addresses the heterogeneous graphs, where the proposed network is capable of generating new graph structures from the original heterogeneous graph by defining meta-paths with arbitrary edge types.  
In~\cite{dwivedi2020generalization}, a generalized graph form of transformers is proposed with the edge features addressed.

\textbf{Camera Pose Regression Networks.} It was not until recently that research interests focus on incorporating deep neural networks into SfM pipelines and camera pose regression tasks~\cite{garg2016unsupervised, kendall2015posenet, tang2018ba, balntas2018relocnet, ding2019camnet, vijayanarasimhan2017sfm, klodt2018supervising, xue2019local}. As one of the earliest work adopting neural networks for camera pose regression, the deep convolutional neural network pose regressor proposed in~\cite{kendall2015posenet} is trained according to a loss function embedding the absolute camera pose prediction error. While~\cite{kendall2015posenet} pioneers in fusing the power of neural networks into pose regression frameworks, it does not take the intra-frame constraints or connectivity of the view-graph into optimization and thus barely over-performs conventional counterparts on the accuracy, as improved later in~\cite{sattler2019understanding, xue2020learning, clark2017vidloc}. Other work exploits the algebraic or geometric relations among the given sequential images and train the networks to predict to locate the images~\cite{brahmbhatt2018geometry, clark2017vidloc, vijayanarasimhan2017sfm, valada2018deep}, among which~\cite{clark2017vidloc} leverages temporal consistency of the sequential images by equipping bi-directional LSTMs with a CNN-RNN model such that temporal regularity can provide more pose information in the regression. The approach in~\cite{brahmbhatt2018geometry} trains DNNs model with the pair-wise geometric constraints between frames, by leveraging additional measurements from IMU and GPS. 

Recent work~\cite{xue2020learning} is the first study to leverage GNNs in a full absolute camera pose regression framework, where the authors model the view-graph with nodes fused with image features extracted by CNNs. Then the CNNs and GNNs are executed iteratively to handle the massage passing for redundant edge removal.

\section{Problem Statement}
\label{sec:problem}
Given a set of 2D image frames representing a 3D scene, in conventional SfM pipelines, {\it camera pose estimation} or {\it camera relocalization} seeks a consistent set of optimized camera rigid motions, aiming to recover the locations and orientations of the camera aligned with the scene coordinate.
Formally, let $\mathbf{R}_i \in \mathbb{SO}(3)$ and $\mathbf{t}_i\in \mathbb{R}^3$ denote the camera orientation and the camera translation for the $i^\text{th}$ image frame respectively, then the absolute camera pose is denoted by $\mathcal{T}_i = [\mathbf{R}_i | \mathbf{t}_i]$.
Accordingly, let $\mathcal{T}_{ij}$ denote the relative camera motion between the $i^\text{th}$ and $j^\text{th}$ image frames, a conventional formulation of the camera pose estimation is to solve the following objective function,
\begin{equation}
    \label{eq:ra}
    \argmin _{\mathbf{R}_i, \mathbf{R}_j } \sum_{(i, j)} \rho \big(d({\mathbf{R}}_{ij}, \mathbf{R}_j \mathbf{R}_i^{-1})\big),
\end{equation}
where $\rho(\cdot)$ is a robust cost function and $d(\cdot, \cdot)$ is the distance metric. That is, given the camera relative orientations $\{\mathbf{R}_{ij}\}$, the optimization process involves minimizing a cost function that penalizes the discrepancy between the camera relative orientations achieved from image retrieval and those inferred from the solved absolute camera poses. We argue that low costs in Eq.~\ref{eq:ra} indicate high global consistency of the solution set and fuse it into the loss function as the consistency loss. Additionally, given the ground truth camera poses $\overline{\mathcal{T}_i} = \big[\overline{\mathbf{R}_i} | \overline{\mathbf{t}_i}\big]$, the objective function is then
\begin{equation}
    \label{eq:wholera}
    \argmin _{\mathbf{R}_i, \mathbf{R}_j } \sum_{(i, j)} \rho \big(d_{\mathbf{R}}({\mathbf{R}}_{ij}, \mathbf{R}_j \mathbf{R}_i^{-1})\big) + \argmin _{\mathbf{R}_i} \sum_{i} \rho' \big(d_{\mathbf{R}}({\mathbf{R}}_{i}, \overline{\mathbf{R}_i} )\big) + \argmin _{\mathbf{t}_i} \sum_{i} \rho'' \big(d_{\mathbf{t}}({\mathbf{t}}_{i}, \overline{\mathbf{t}_i} )\big),
\end{equation}
where $\rho'$ and $\rho''$ are robust cost functions, $d_{\mathbf{R}}:\mathbb{SO}(3) \times \mathbb{SO}(3) \rightarrow \mathbb{R}_{+}$ and $d_{\mathbf{t}} : \mathbb{R}^3 \times \mathbb{R}^3 \rightarrow \mathbb{R}_{+}$ are the distance metric for rotations and translations respetively. More details on the loss function formulation are given in \S\ref{sec:graphloss}.

In the design of our proposed network, we model the multi-view camera relocalization problem as graphs and embed the 2D image features and the camera absolute pose $\mathcal{T}_i$ as the corresponding latent node information, whereas the inter-frame camera relative motions $\mathcal{T}_{ij}$ are encoded as the edge attributes, as introduced in \S\ref{sec:graph}.  

\section{TransCamP Architecture}
\label{sec:architecture}
In this section we detail the network architecture of the proposed TransCamP. First we provide the architecture overview in \S\ref{sec:overview}, followed by the elaboration of feature embedding and graph embedding in \S\ref{sec:graph}. We then emphasize the structure of the spatiotemporal graph transformer layers in \S\ref{sec:networklayer}, followed by the graph update and the proposed graph loss function illustrated in \S\ref{sec:graphloss}.
\subsection{Architecture Overview}
\label{sec:overview}
As shown in Fig.~\ref{fig:pipeline}, the proposed network takes images as RGB images as input, the images are first fed into a pre-trained CNN-type feature network, then the output feature maps are embedded in a initial view-graph such that the nodes encode the visual information of the images and the edges encode inter-frame correlations. Additionally, the local feature matching information and the aggregated image matching score are arranged into a tensorized adjacency matrix. 

After ensembling the images into a graph, the adjacency tensor and the hidden node features are first passed into MPNN layers such that, for each node, the neighboring node features are aggregated efficiently with the implicit attention information embedded in the adjacency tensor. Then the aggregated node features are fed into graph transformer encoder layers, where the self attention mechanism are equipped with edge features such that the camera relative transformations encoded on the edges can be exploited to generate the attention weights. Additionally, the temporal transformer encoder layers capture the self-attention for the sequential input. The global camera poses, as node attributes, are updated through the network and are embedded in the final output.  
\subsection{Graph Embedding}
\label{sec:graph}
We propose to model the input images, the corresponding camera poses and the pair-wise camera transformations into a graph based on the construction of conventional pose graph, \ie, each node represents an image frame and the edges connecting two nodes represent the inter-frame image relations. In detail, consider a graph $\mathcal{G} = (\mathcal{V}, \mathcal{E})$ where $\mathcal{V} = \{v_i\}$ denotes the set of the images and $\mathcal{E} = \{(i,j) | v_i, v_j \in \mathcal{V}\}$ represents the pair-wise feature-base connectivity between frames. Additionally, let $\mathcal{A}_{\mathcal{G}}$ denote the adjacency matrix of $\mathcal{G}$ such that $\mathcal{A}_{\mathcal{G}}(i,j) = 0$ if $(i,j) \not\in \mathcal{E}$ and vice versa. For simplicity of notation, we will use $\mathcal{A}$ for $\mathcal{A}_{\mathcal{G}}$ in the following discussion.

\textbf{Node Attributes.}
Consider an image $\mathbf{I}_i$, let $\mathbf{x}_i$ denote the feature vector as the output of the CNN-type feature sub-network, and denote $\mathbf{p}_i\in \mathbb{R}^7$ as the camera absolute pose vector, where $\mathbf{p}_i$ consists of the 4-dimensional quaternion $\omega _i$ representing the camera orientation and the 3-dimensional $t_i$ representing the camera translation. That is, the vector embedding of each node $v_i$ contains the information part which encodes the image latent feature and the learning part which embeds the camera pose. It is noteworthy to mention that, in contrast with NLP tasks where the word positions or text orders are crucial, the camera absolute poses are invariant to node positions as we leverage the graph structure to model the problem. Therefore we skip the positional encoding in the original Transformer model~\cite{vaswani2017attention} and embed the `relative position' or `relative distance' as the image matching vector into the tensorized adjacency matrix.

\begin{wrapfigure}{r}{0.5\textwidth}
\vspace{-5mm}
\begin{center}
    \includegraphics[width = 0.48\textwidth]{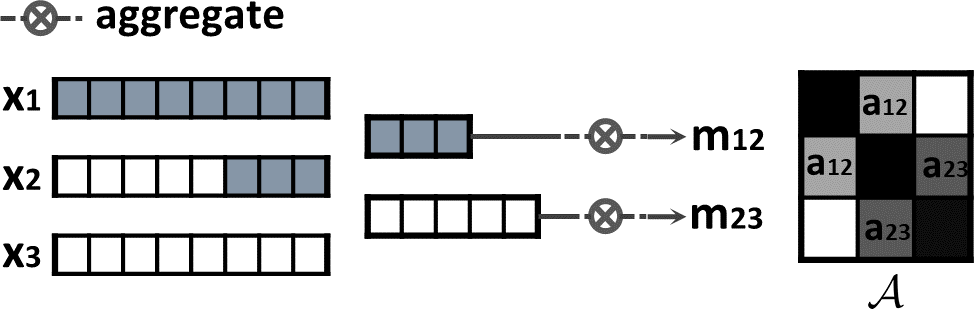}
\end{center}
\vspace{-2mm}
\caption{Each element $a_{ij}$ of the tensorized adjacency matrix $\mathcal{A}$ embeds the feature correspondences and the normalized aggregated value. $m_{ij}=0$ if there exists none co-visible feature between image $i$ and image $j$. Note that $\mathcal{A}$ is symmetric.}
\label{fig:adjmat}
\vspace{-2mm}
\end{wrapfigure}
\textbf{Adjacency Tensor.} Let $a_{ij}$ be the element at $(i,j)$ of the adjacency matrix with self-connections $\mathcal{A}$, by convention $a_{ij} = 1$ if there exists an edge connecting $v_i$ and $v_j$ and $a_{ij} = 0$ otherwise. To capture and maintain the pair-wise relation, we introduce the adjacency tensor where $a_{ij}$ represents the vector feature correspondence index between the $i^{\text{th}}$ and $j^{\text{th}}$ image frames.

Specifically, consider $a_{ij} \in \mathcal{A}$ and let $\mathbf{x}_i$ and $\mathbf{x}_j$ be the corresponding feature vectors and assume that there exists some feature correspondences between image $i$ and image $j$. Then $a_{ij}^k$, \ie, the $k^{\text{th}}$ element of $a_{ij}$ portraying the $k^{\text{th}}$ feature correspondence, is a tuple with the feature index in $\mathbf{x}_i$ and $\mathbf{x}_j$ respectively. That is, $\mathbf{x}_i(a_{ij}^k(1)) = \mathbf{x}_j (a_{ij}^k(2))$. Additionally, each vector $a_{ij}$ is aggregated into an initial meta-feature $\mathbf{m}_{ij}$ as the normalized feature correspondence score with range $[0, 1]$, which measures the edge credibility evaluation and the image matching result between the two connected nodes. The tensorized adjacency matrix encodes pixel-wise and image-wise correspondence, depicting the edge weights and is updated through the network while interacting spatiotemporally with the whole evolving graph. Illustration of the adjacency tensor is given in Fig.~\ref{fig:adjmat}.

\textbf{Edge Attributes.}
Similar with the 7-dimensional pose feature embedded on the nodes, the camera relative transformation is encoded on the edge connecting $v_i$ and $v_j$ as $\mathbf{p}_{ij} = <\omega _{ij}, t_{ij}>$. During the graph embedding, only nodes with matched features are connected with edges with initialized edge feature (unit quaternion translation and zero vector translation).
 In our modeling of the graph we consider the edge features as node-symmetric according to the nature of pose graph construction. As we aim to keep the graph lightweight, the edges do not contain any low-level correspondence information between the connected nodes. Instead, the inter-node dependency is implicitly arranged into the adjacency tensor $\mathcal{A}$.

\subsection{Graph Transformer Layer}
\label{sec:networklayer}
Now we have constructed the graph embedding the node and edge features as the input into the graph transformer layer. Our proposed network adopts the encoder layer structure in the original transformer~\cite{vaswani2017attention} and transforms the initial source graph to the target graph with evolved structural edge information and derived pose values on the nodes. Specifically, the graph transformer layer exploits the multi-head attention mechanism to generate the sptiotemporal relation between nodes, such that 1) the edges connecting two nodes where high amounts of common features (pair-wise co-visible visual features) are equipped with high attention weights and 2) the edges carrying abundant or noisy image matching yield low attention weights or get removed from the graph. The emerging adjacency tensor progressively interacts with the whole graph and propagates the update over the nodes and the edges.

\begin{wrapfigure}{r}{0.45\textwidth}
\vspace{-6mm}
\begin{center}
    \includegraphics[width = 0.40\textwidth]{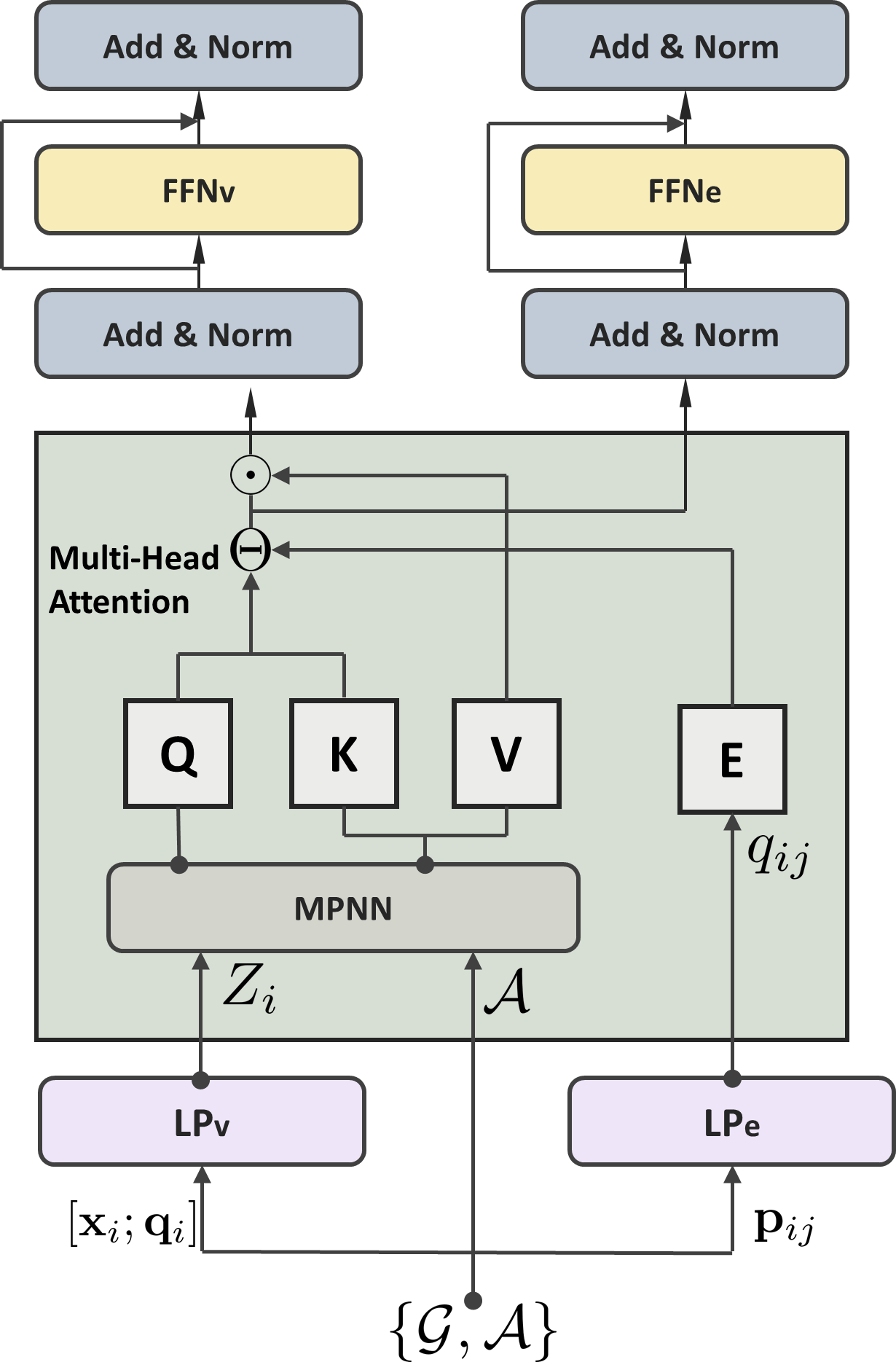}
\end{center}
\vspace{-2mm}
\caption{Illustration of the graph transformer encoder layer structure.}
\label{fig:layer}
\end{wrapfigure}

\textbf{Message Passing.} Before passing the graph into the graph transformer encoder layer, the neighboring node features are aggregated along with the adjacency tensor for each node. Specifically, consider the graph at the $l^{\text th}$ layer and let $Z^l$ denote the hidden feature tensor of the nodes, let $\mathcal{A}^l$ denote the adjacency tensor. Then after the message passing layers the node tensor is thus
\begin{equation}
    \label{eq:msgnode}
    \hat{Z}^l = \doubleplus [\phi (\mathcal{A}^l, Z^l) \otimes Z^l],
\end{equation}
where $\doubleplus$ denotes the concatenation operation, $\phi(,)$ denotes the message aggregation, $\otimes$ denotes the tensor product. The adjacency tensor is exploited here instead of the edges as $\mathcal{A}$ has collected the local attention information such that the message passing is more efficient.

\textbf{Graph Transformer Encoder Layer.} We leverage the multi-head self attention mechanism in the graph transformer encoder layer with edge features. Borrowing notations from the original transformer network, let $Q_k^l, K_k^l, V_k^l \in \mathbb{R}^{d_k \times d}$, where $k = 1$ to $N$ is the number of the attention heads, $d_k$ denotes the query dimension. Consider the attention weight for the $k^\text{th}$ head on the edge connecting the source node $i$ and the target node $j$, that is
\begin{equation}
    \label{eq:nodeatt}
    w_{ij} = \text{softmax}_j (Q_k^l \hat{Z}_i^l \odot K_k^l \hat{Z}_j^l ).
\end{equation}
Following~\cite{dwivedi2020generalization}, we add the edge features into generating the attention. Let $E_k^l$ be in the same dimension space with $Q_k^l, K_k^l, V_k^l$ and let $q_{ij}$ denote the hidden edge features, then the attention weight with edge feature is thus
\begin{equation}
    \label{eq:newnodeatt}
    w_{ij}^e =\text{softmax}_j \Theta(Q_k^l \hat{Z}_i^l, K_k^l \hat{Z}_j^l, E_k^l q_{ij}^l ),
\end{equation}
where $\Theta$ denotes the dot product operation. Then the update function for nodes and edges are thus
\begin{align}
    \label{eq:update}
    Z_i^{l+1} &= \doubleplus_k(w_{ij}^e V_k^l \hat{Z}_j^l) \otimes O_Z^l,\\
    q_{ij}^{l+1} &= \doubleplus_k (w_{ij}^e) \otimes O_e^l,
\end{align}
where $O_Z^l, O_e^l \in \mathbb{R}^{d\times d}$, $d$ is the dimension of the hidden feature space of the nodes and edges. Illustration is given in Fig.~\ref{fig:layer}.

\textbf{Temporal Transformer Encoder Layer.} The temporal inter-frame relation contains high amounts of useful information especially when the input is sequential images or video clips. In the proposed network we address the temporal dependencies for consecutive camera relocalization tasks by equipping the network with an optional temporal transformer encoder layer. The temporal transformer encoder layer exploits the standard Transformer network structure, takes the graph embedding as input and generates intra-graph temporal dependencies between nodes by constructing temporal attention.  

\subsection{Graph Loss Function}
 \label{sec:graphloss}
 The TransCamP is trained end-to-end, guided by the joint loss function representing both the graph consistency and the accuracy of the predicted camera poses. Recall the objection function Eq.~\ref{eq:wholera}, the loss function is thus assembled as
 \begin{align}
     \nonumber \mathcal{L} &= \alpha \sum_{i,j} \rho (d_{\mathbf{R}} (\omega_{ij}, {\omega}_j {\omega}_i^{-1})) + \alpha ' \sum_{i,j} \rho ' (d_{\mathbf{t}} (t_{ij}, d_{\mathbf{t}}(t_i, t_j)))\\
     &+ 
 \beta \sum_{i} \rho (d_{\mathbf{R}} (\omega_{i}, \overline{{\omega}_i})) + \beta ' \sum_{i} \rho ' (d_{\mathbf{t}} (t_{i}, \overline{t_i})),
 \end{align}
 where $\alpha,\alpha', \beta, \beta'\in \mathbb{R}$ are the loss parameters, $\overline{\omega_i}, \overline{t_i}$ are the ground truth camera orientations and translations. The graph loss function can be seen as a joint optimization regarding both the graph consistency and the prediction accuracy.
\section{Experimental Results}
\label{sec:experiment}
The proposed network is evaluated on three public benchmarks: 7Scenes~\cite{shotton2013scene}, the Cambridge dataset~\cite{kendall2015posenet} and the Oxford Robotcar dataset~\cite{maddern2017oxford}. We first elaborate the datasets, metrics, baselines and implementation details we conduct the experiments with (\S\ref{sec:implem}), followed by the evaluation results (\S\ref{sec:evalresults}), we then conduct the ablation study on the spatiotemporal mechanism of the proposed network (\S\ref{sec:ablation}) and discuss the limitations (\S\ref{sec:limit}).

\subsection{Experiment Setting}
\label{sec:implem}
\textbf{Implementation Details.} The proposed network is implemented in PyTorch on a machine with Intel(R) i7-7700 3.6GHz processors with 8 threads and 64GB memory and single Nvidia GeForce 1080 GPU with 8GB memory. 
For training we adopt standard SGD optimizer with no dropout, the learning rate is annealed geometrically starting at 1e-3 and decreases to 1e-5. 

We adopt ResNet~\cite{he2016deep} pretrained on ImageNet~\cite{deng2009imagenet} for the feature handling. The input RGB images are scaled to 341$\times$ 256 pixels, normalized by the subtraction of mean pixel values. The proposed network is trained end-to-end on ScanNet~\cite{dai2017scannet}, a RGB-D video sequence dataset which contains 2.5million views in over 1500 indoor scans, we only use the RGB monocular images and the ground truth camera pose values are given by~\cite{dai2017bundlefusion}.   

\textbf{Datasets and Metrics.} We conduct extensive experiments on datasets with different scales and report the median errors of camera orientation ($^\circ$) and translation (m). The {\it 7Scenes dataset}~\cite{shotton2013scene} consists of RGB-D video sequences covering seven small indoor scenes, captured by hand-held Kinect camera. In some of the scenes, many texture-less surfaces and repetitive patterns are present, thus making the dataset challenging in spite of its relatively small size containing less than 10K images. The {\it Cambridge dataset} is a large-scale dataset containing six outdoor scene scans outside the Cambridge University, the dataset consists of around 12K images and the corresponding camera pose ground truth.  
The {\it Oxford RobotCar dataset} contains image sequences taken through driving in Oxford with different weathers, traffic conditions and lighting, the total trajectory is over 10km and is very challenging for camera relocalization. Following~\cite{brahmbhatt2018geometry, xue2019local, xue2020learning}, we conduct experiments on the LOOP route (1120m) and FULL route (9562m) to evaluate the performance of the proposed network on long consecutive sequences.  

\textbf{Baselines.} The proposed network is evaluated against recent state-of-the-art camera relocalization networks, including single image-based absolute camera pose regression network PoseNet and its variants~\cite{kendall2015posenet, kendall2016modelling, kendall2017geometric, walch2017image} among which,  LSTM+Pose~\cite{walch2017image} along with MapNet and its variants~\cite{brahmbhatt2018geometry}, LsG~\cite{xue2019local} and VidLoc~\cite{clark2017vidloc} have utilized temporal inter-frame relations in the network. CNN+GNN~\cite{xue2020learning} models the multi-view camera pose estimation with a graph and leverages GNNs on the task. Other approaches include RelocNet~\cite{balntas2018relocnet}, Hourglass~\cite{melekhov2017image} and BranchNet~\cite{wu2017delving}. 
\subsection{Performance Evaluation}
\label{sec:evalresults}
\textbf{7Scenes.} We first evaluate TransCamP on the 7 Scenes dataset against recent state-of-the-art approaches, the experiment results are given in Table.~\ref{table:7scenes}. It can be observed that our proposed network overperforms the other approaches on most of the scenes. Among the approaches, LsG~\cite{xue2019local}, MapNet~\cite{brahmbhatt2018geometry} and VidLoc~\cite{clark2017vidloc} rely heavily on the temporal information of the input, \ie, the approaches can handle consecutive sequences more efficiently but tend to lose the spatial inter-frame correlation especially for large-scale datasets or over long camera trajectories. Additionally, PoseNet~\cite{kendall2015posenet} and its variants conduct absolute pose regression from single images, such that the networks perform poorly on the scene where repetitive patterns or texture-less surfaces are present. 

Similar to our proposed network, CNN+GNN~\cite{xue2020learning} leverages graphs to model the multi-view camera relocalization with message passing among the image-embedded nodes. However, the network does not exploit temporal information in sequential images, and enforces a maximum value of neighbors of each node. As a result, it tends to miss the temporal correlation for consecutive frames or discard useful inter-frame spatial correlation.   

\begin{table*}[t!]
\midsepremove
\begin{center}
    \caption{Experiment results on the 7Scenes Dataset~\cite{shotton2013scene}. Results are cited directly, the best results are \textbf{{highlighted}}.}
    \label{table:7scenes}
    \resizebox{\linewidth}{!}{%
    \begin{tabular}{llllccccccc|c}
    \toprule[0.5mm]
    \multicolumn{4}{l}{\makecell{\textbf{Scene}\\ Scene scale}} & {\makecell{\textbf{Chess}\\{3 x 2m$^2$ }}}& {\makecell{\textbf{Fire}\\{2.5 x 1m$^2$ }}}& {\makecell{\textbf{Heads}\\{2 x 0.5m$^2$ }}}& {\makecell{\textbf{Office}\\{2.5 x 2m$^2$ }}}& {\makecell{\textbf{Pumpkin}\\{2.5 x 2m$^2$ }}}& {\makecell{\textbf{Kitchen}\\{4 x 3m$^2$ }}}& {\makecell{\textbf{Stairs}\\{2.5 x 2m$^2$ }}}& {\textbf{Avg.}} \\ \midrule[0.3mm]
          \multicolumn{4}{l}{RelocNet~\cite{balntas2018relocnet}} & 0.12m, 4.14$^\circ$ & 0.26m, 10.4$^\circ$ & 0.14m, 10.5$^\circ$ & 0.18m, 5.32$^\circ$ & 0.26m, 4.17$^\circ$ & 0.23m, 5.08$^\circ$ & 0.28m, 7.53$^\circ$ &  0.21m, 6.73$^\circ$\\ 
          \multicolumn{4}{l}{LsG~\cite{xue2019local}} & 0.09m, 3.28$^\circ$ & 0.26m, 10.92$^\circ$& 0.17m, 12.70$^\circ$ & 0.18m, 5.45$^\circ$ & 0.20m, 3.69$^\circ$ & 0.23m, 4.92$^\circ$ & 0.23m, 11.3$^\circ$ & 0.19m, 7.47$^\circ$\\ 
           \multicolumn{4}{l}{MapNet~\cite{brahmbhatt2018geometry}} & 0.08m, 3.25$^\circ$ & 0.27m, 11.69$^\circ$ & 0.18m, 13.25$^\circ$ & 0.17m, 5.15$^\circ$ & 0.22m, 4.02$^\circ$ & 0.23m, 4.93$^\circ$ & 0.30m, 12.08$^\circ$ & 0.21m, 7.77$^\circ$\\
               \multicolumn{4}{l}{MapNet+~\cite{brahmbhatt2018geometry}} & 0.10m, 3.17$^\circ$ & \textbf{0.20m}, 9.04$^\circ$ & 0.13m, 11.13$^\circ$ & 0.18m, 5.38$^\circ$ & 0.19m, 3.92$^\circ$ & 0.20m, 5.01$^\circ$ & 0.30m, 13.37$^\circ$ & 0.19m, 7.29$^\circ$\\
          \multicolumn{4}{l}{MapNet(pgo)~\cite{brahmbhatt2018geometry}} & 0.09m, 3.24$^\circ$ & \textbf{0.20m}, 9.29$^\circ$ & 0.12m, \textbf{8.45$^\circ$} & 0.19m, 5.42$^\circ$ & 0.19m, 3.96$^\circ$ & 0.20m, 4.94$^\circ$ & 0.27m, 10.57$^\circ$ & 0.18m, 6.55$^\circ$\\
           \multicolumn{4}{l}{PoseNet15~\cite{kendall2015posenet}} & 0.32m, 8.12$^\circ$ & 0.47m, 14.4$^\circ$ & 0.29m, 12.0$^\circ$ & 0.48m, 7.68$^\circ$ & 0.47m, 8.42$^\circ$ & 0.59m, 8.64$^\circ$ & 0.47m, 13.8$^\circ$ & 0.44m, 10.4$^\circ$\\
           \multicolumn{4}{l}{PoseNet16~\cite{kendall2016modelling}} & 0.37m, 7.24$^\circ$ & 0.43m, 13.7$^\circ$ & 0.31m, 12.0$^\circ$ & 0.48m, 8.04$^\circ$ & 0.61m, 7.08$^\circ$ & 0.58m, 7.54$^\circ$ & 0.48m, 13.1$^\circ$ & 0.47m, 9.81$^\circ$\\
           \multicolumn{4}{l}{PoseNet17~\cite{kendall2017geometric}} & 0.14m, 4.50$^\circ$ & 0.27m, 11.80$^\circ$ & 0.18m, 12.10$^\circ$ & 0.20m, 5.77$^\circ$ & 0.25m, 4.82$^\circ$ & 0.24m, 5.52$^\circ$ & 0.37m, 10.60$^\circ$ & 0.24m, 7.87$^\circ$\\
           \multicolumn{4}{l}{PoseNet17+~\cite{kendall2017geometric}} & 0.13m, 4.48$^\circ$ & 0.27m, 11.30$^\circ$ & 0.17m, 13.00$^\circ$ & 0.19m, 5.55$^\circ$ & 0.26m, 4.75$^\circ$ & 0.23m, 5.35$^\circ$ & 0.35m, 12.40$^\circ$ & 0.23m, 8.12$^\circ$\\
          \multicolumn{4}{l}{LSTM+Pose\cite{walch2017image}} & 0.24m, 5.77$^\circ$ & 0.34m, 11.9$^\circ$ & 0.21m, 13.7$^\circ$ & 0.30m, 8.08$^\circ$ & 0.33m, 7.00$^\circ$ & 0.37m, 8.83$^\circ$ & 0.40m, 13.7$^\circ$ & 0.31m, 9.85$^\circ$\\
           \multicolumn{4}{l}{Hourglass~\cite{melekhov2017image}} & 0.15m, 6.17$^\circ$ & 0.27m, 10.84$^\circ$ & 0.19m, 11.63$^\circ$ & 0.21m, 8.48$^\circ$ & 0.25m, 7.01$^\circ$ & 0.27m, 10.15$^\circ$ & 0.29m, 12.46$^\circ$ & 0.23m, 9.53$^\circ$\\
           \multicolumn{4}{l}{BranchNet~\cite{wu2017delving}} & 0.18m, 5.17$^\circ$ & 0.34m, 8.99$^\circ$ & 0.20m, 14.15$^\circ$ & 0.30m, 7.05$^\circ$ & 0.27m, 5.10$^\circ$ & 0.33m, 7.40$^\circ$ & 0.38m, 10.26$^\circ$ & 0.29m, 8.30$^\circ$\\
           \multicolumn{4}{l}{VidLoc~\cite{clark2017vidloc}} & 0.18m, - & 0.26m, - & 0.14m, - & 0.26m, - & 0.36m, - & 0.31m, - & 0.26m, - & 0.25m, -\\
           \multicolumn{4}{l}{CNN+GNN~\cite{xue2020learning}} & \textbf{0.08m}, 2.82$^\circ$ & 0.26m, 8.94$^\circ$ & 0.17m, 11.41$^\circ$ & 0.18m, 5.08$^\circ$ & \textbf{0.15m}, 2.77$^\circ$ & 0.25m, 4.48$^\circ$ & \textbf{0.23m}, \textbf{8.78$^\circ$} & 0.19m, 6.33$^\circ$\\ \midrule[0.3mm]
          \multicolumn{4}{l}{\textbf{Ours}} & \textbf{0.08m, 1.97$^\circ$} & 0.27m, \textbf{8.26$^\circ$} & \textbf{0.12m}, 9.66$^\circ$ & \textbf{0.12m, 2.37$^\circ$} & 0.16m, \textbf{2.49$^\circ$} & \textbf{0.19m, 2.64$^\circ$} & 0.28m, 9.01$^\circ$ & \textbf{0.17m, 5.2$^\circ$}\\ \toprule[0.5mm]
    \end{tabular}
    }
\end{center}
\end{table*}

\begin{table*}[t!]
\midsepremove
\begin{center}
    \caption{Experiment results on the Cambridge Dataset~\cite{kendall2015posenet}. Evaluation with MapNet~\cite{brahmbhatt2018geometry} is cited from~\cite{sattler2019understanding}, other results are cited directly. The average is taken on the first four datasets. The best results are \textbf{{highlighted}}.}
    \label{table:cambridge}
    \resizebox{\linewidth}{!}{%
    \begin{tabular}{llllcccccc|c}
    \toprule[0.5mm]
    \multicolumn{4}{l}{\makecell{\textbf{Scene}\\ Scene scale}} & {\makecell{\textbf{College}\\{5.6x10$^3$ m$^2$}}}& {\makecell{\textbf{Shop}\\{8.8x10$^3$ m$^2$}}}& {\makecell{\textbf{Church}\\{4.8x10$^3$ m$^2$}}}& {\makecell{\textbf{Hospital}\\{2.0x10$^3$ m$^2$}}}& {\makecell{\textbf{Court}\\{8.0x10$^3$ m$^2$}}}& {\makecell{\textbf{Street}\\{5.0x10$^3$ m$^2$}}}& {\textbf{Avg.}} \\ \midrule[0.3mm]

           \multicolumn{4}{l}{MapNet~\cite{brahmbhatt2018geometry}} & 1.07m, 1.89$^\circ$ & 1.49m, 4.22$^\circ$ & 2.00m, 4.53$^\circ$ & 1.94m, 3.91$^\circ$ & 7.85m, 3.76$^\circ$ & 22.23m, 27.55$^\circ$ & 1.63m, 3.64$^\circ$\\
               
           \multicolumn{4}{l}{PoseNet15~\cite{kendall2015posenet}} & 1.66m, 4.86$^\circ$ & 1.41m, 7.18$^\circ$ & 2.45m, 7.96$^\circ$ & 2.62m, 4.90$^\circ$ & - & - & 2.04m, 6.23$^\circ$\\
           \multicolumn{4}{l}{PoseNet16~\cite{kendall2016modelling}} & 1.74m, 4.06$^\circ$ & 1.25m, 7.54$^\circ$ & 2.11m, 8.38$^\circ$ & 2.57m, 5.14$^\circ$ & - & - & 1.92m, 6.28$^\circ$\\
           \multicolumn{4}{l}{LSTM+Pose~\cite{walch2017image}} & 0.99m, 3.65$^\circ$ & 1.18m, 7.44$^\circ$ & \textbf{1.52m}, 6.68$^\circ$ & 1.51m, 4.29$^\circ$ & - & - & 1.30m, 5.52$^\circ$\\
           \multicolumn{4}{l}{PoseNet17~\cite{kendall2017geometric}} & 0.99m, 1.06$^\circ$ & 1.05m, 3.97$^\circ$ & 1.49m, 3.43$^\circ$ & 2.17m, 2.94$^\circ$ & 7.00m, 3.65$^\circ$ & 20.70m, 25.70$^\circ$ & 1.43m, 2.85$^\circ$\\
           \multicolumn{4}{l}{PoseNet17+~\cite{kendall2017geometric}} & 0.88m, 1.04$^\circ$ & 0.88m, 3.78$^\circ$ & 1.57m, 3.32$^\circ$& 3.20m, 3.29$^\circ$ & 6.83m, 3.47$^\circ$ & 20.30m, 25.50$^\circ$ & 1.63m, 2.86$^\circ$\\
          
           \multicolumn{4}{l}{CNN+GNN~\cite{xue2020learning}} & 0.59m 0.65$^\circ$ & 0.50m, 2.87$^\circ$ & 1.90m, 3.29$^\circ$ & 1.88m, 2.78$^\circ$ & 6.67m, 2.79$^\circ$ & 14.72m, 22.44$^\circ$ & 1.12m, 2.40$^\circ$\\ \midrule[0.3mm]
          \multicolumn{4}{l}{\textbf{Ours}} & \textbf{0.42m, 0.36$^\circ$} & \textbf{0.24m, 1.56$^\circ$} & 1.55m, \textbf{2.56$^\circ$} & \textbf{1.01m, 1.83$^\circ$} & \textbf{4.57m, 2.03$^\circ$} &\textbf{ 9.35m, 17.67$^\circ$} & \textbf{0.81m, 1.58$^\circ$}\\ \toprule[0.5mm]
    \end{tabular}
    }
\end{center}
\end{table*}

\textbf{Cambridge.} We demonstrate the capability to handle large-scale dataset of TransCamP by evaluating the network on the Cambridge dataset, where the proposed network outperforms the baselines on most of the scenes. Among the scenes, `Court' and `Street' are the largest datasets in size and cover long complex trajectories and huge outdoor areas, as challenging to handle with single image-based regression networks like PoseNet15, PoseNet16 and even LSTM+Pose with additional LSTM units, the aforementioned networks have not reported the results on these two datasets. It can be observed that TransCamP demonstrates great improvements over approaches solely relying on temporal relation or spatial relation on datasets with long camera trajectories.     

\begin{wraptable}{r}{0.5\textwidth}
\midsepremove
\begin{center}
    \caption{Experiment results on the Oxford Robotcar Dataset~\cite{maddern2017oxford}. Evaluation with PoseNet~\cite{kendall2015posenet} is cited from ~\cite{brahmbhatt2018geometry}, other results are cited directly, the best results are \textbf{{highlighted}}.}
    \label{table:oxford}
    \resizebox{\linewidth}{!}{%
    \begin{tabular}{lll|cc}
    \toprule[0.5mm]
    \multicolumn{3}{l}{\makecell{\textbf{Scene}\\ Scene scale}} & {\makecell{\textbf{LOOP}\\{1120m }}} & {\makecell{\textbf{FULL}\\{9562m }}}\\
    \midrule[0.5mm]
     \multicolumn{3}{l}{MapNet~\cite{brahmbhatt2018geometry}} & 9.84m, 3.96$^\circ$ & 41.4m, 12.5$^\circ$\\
     \multicolumn{3}{l}{MapNet+~\cite{brahmbhatt2018geometry}} & 8.17m, 2.62$^\circ$ & 30.3m, 7.8$^\circ$\\
     \multicolumn{3}{l}{MapNet(pgo)~\cite{brahmbhatt2018geometry}} & 6.73m, 2.23$^\circ$ & 29.5m, 7.8$^\circ$\\
      \multicolumn{3}{l}{PoseNet~\cite{kendall2015posenet}} & 25.29m, 17.45$^\circ$ & 125.6m, 27.1$^\circ$\\
       \multicolumn{3}{l}{LsG~\cite{xue2019local}} &9.19m, 3.52$^\circ$ & 31.65m, 4.51$^\circ$\\
       \multicolumn{3}{l}{CNN+GNN~\cite{xue2020learning}}& 8.15m, 2.57$^\circ$ & 17.35m, \textbf{3.47$^\circ$}\\
       \midrule[0.3mm]
       \multicolumn{3}{l}{\textbf{Ours}} &\textbf{3.27m, 1.99$^\circ$} & \textbf{12.06m}, 3.66$^\circ$\\
      \toprule[0.5mm]
     
\end{tabular}
    }
\end{center}
\end{wraptable}

\begin{figure*}
\centering
\subfloat[Translation error on LOOP]{
\includegraphics[width =0.43\linewidth]{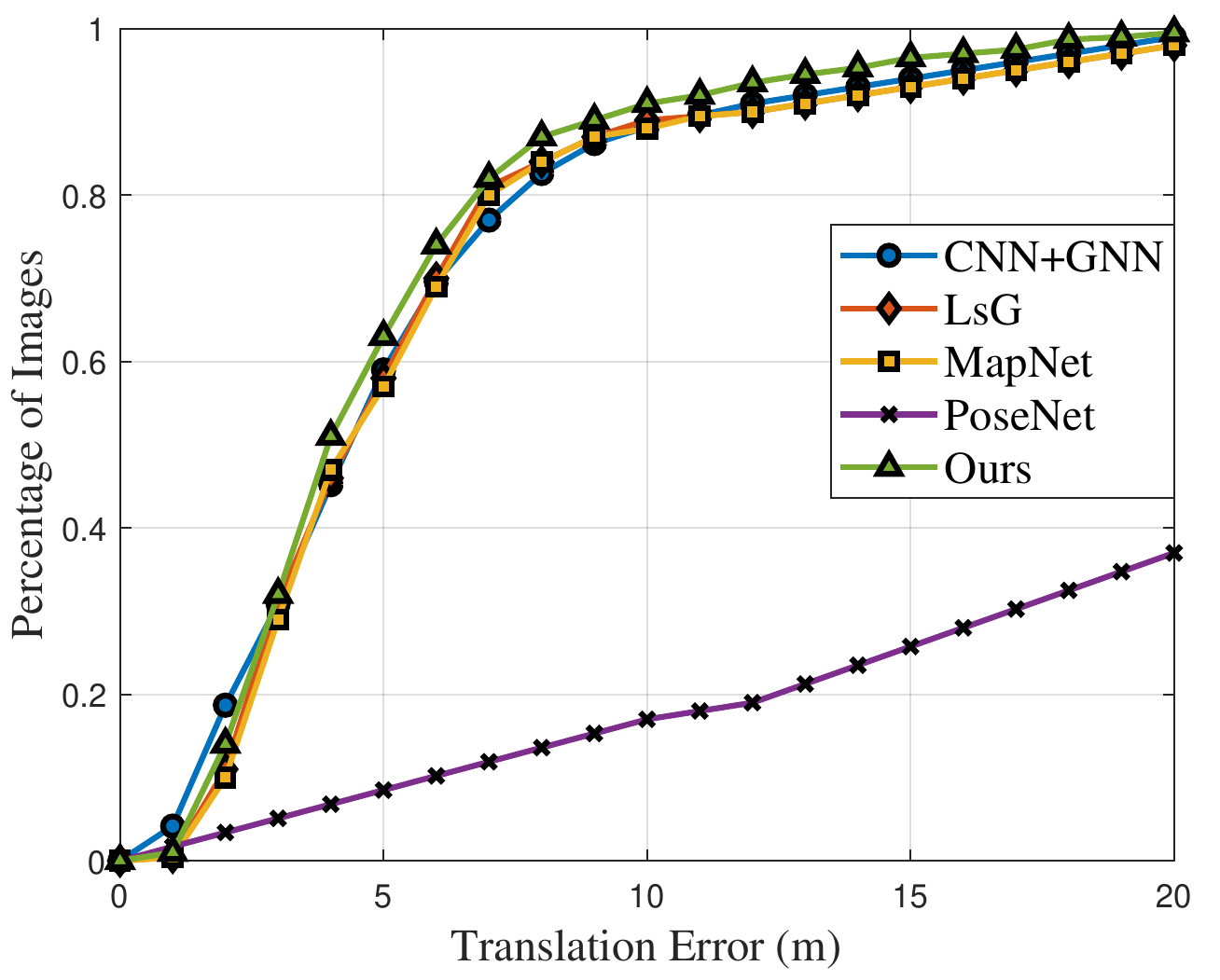}
}
\subfloat[Rotation error on LOOP]{
\includegraphics[width =0.43\linewidth]{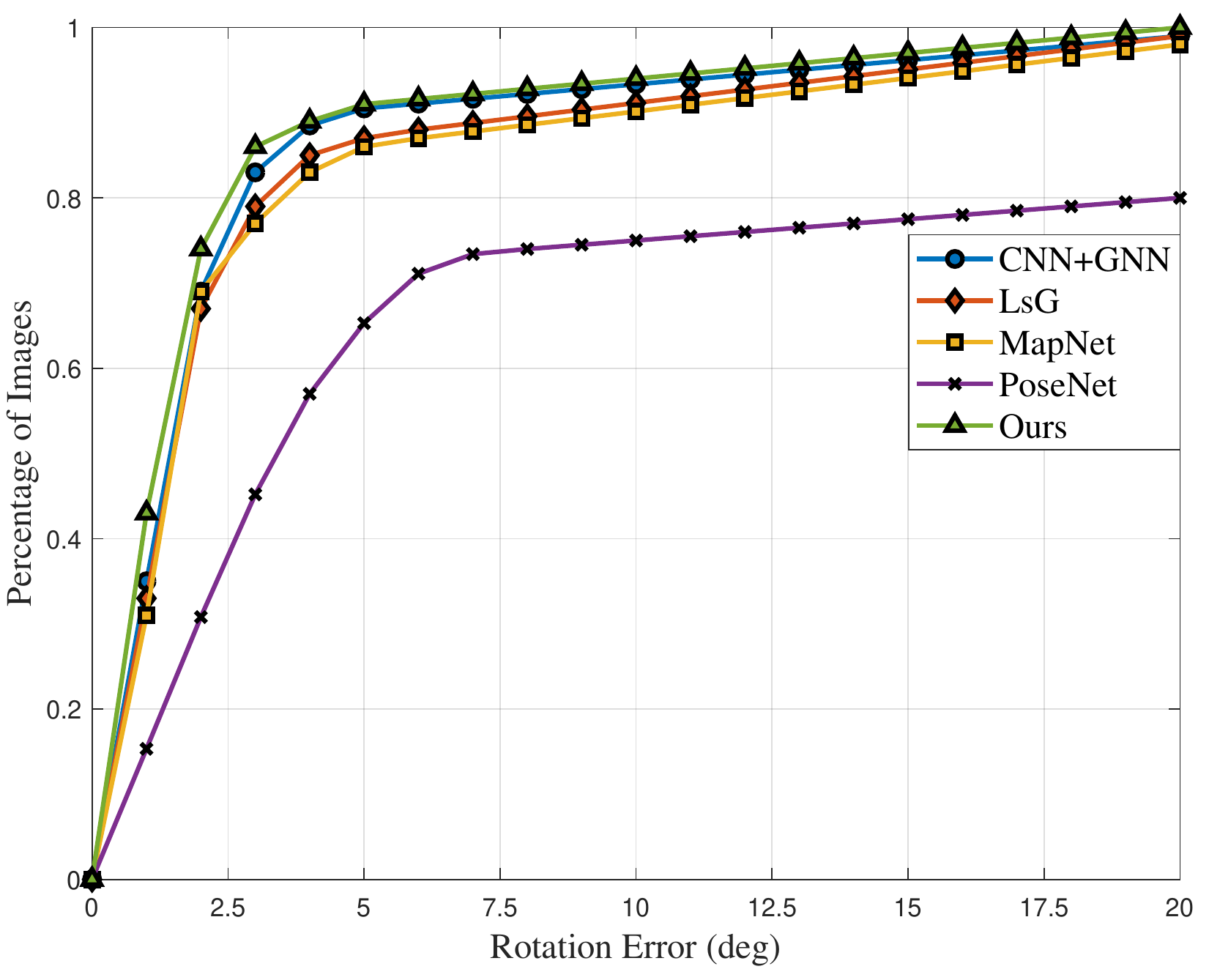}
}

\subfloat[Translation error on FULL]{
\includegraphics[width =0.43\linewidth]{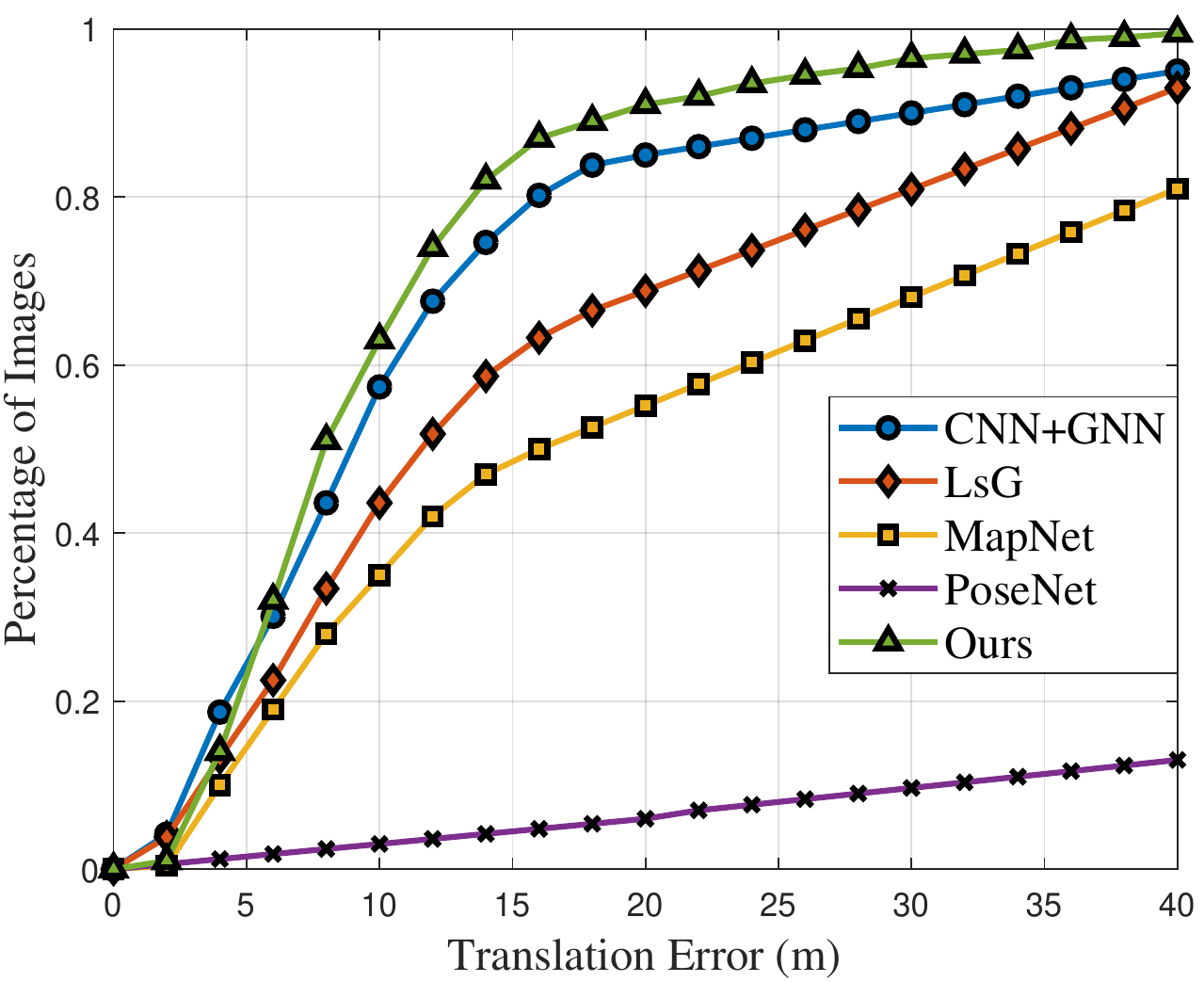}
}
\subfloat[Rotation error on FULL]{
\includegraphics[width =0.43\linewidth]{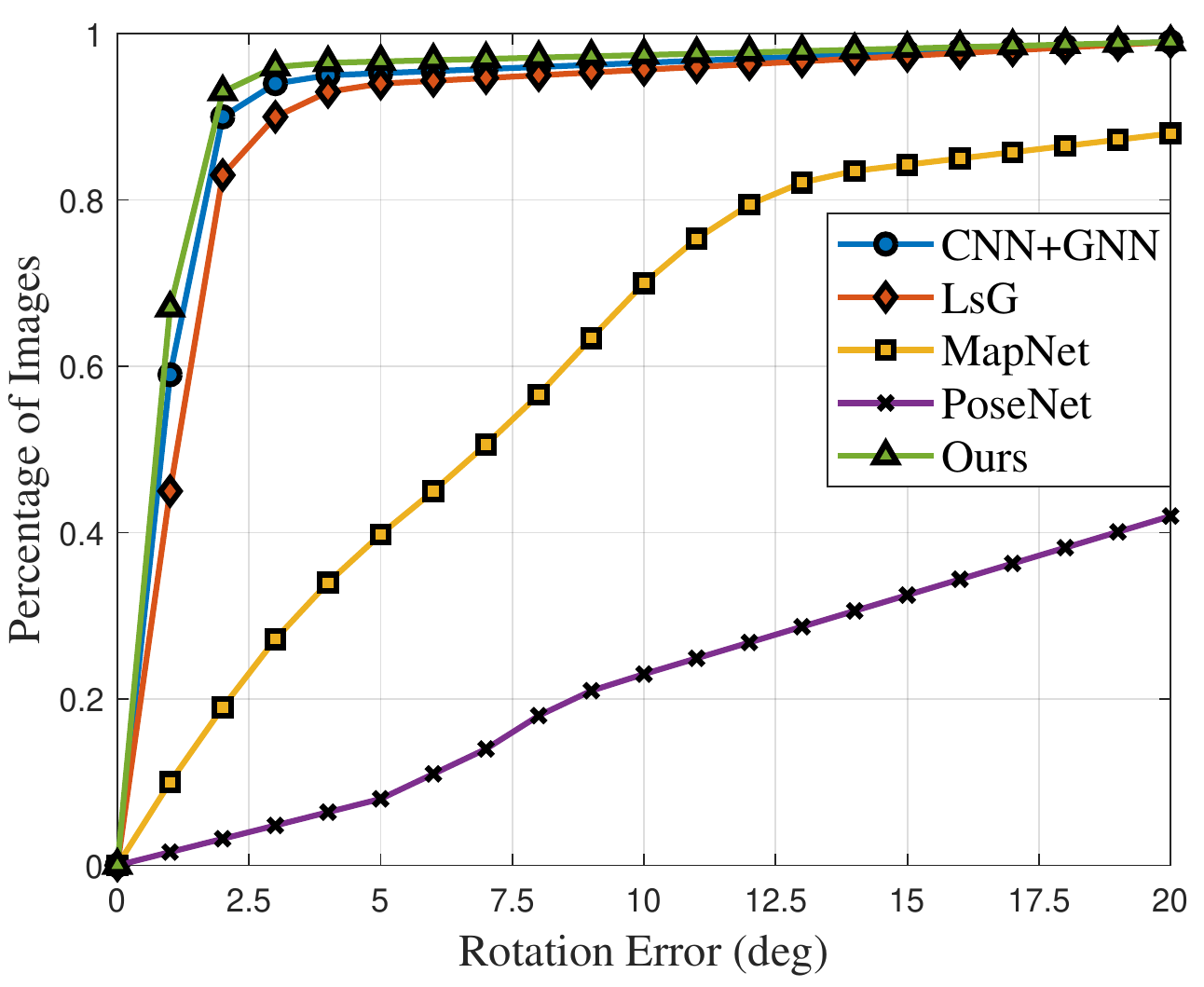}
}
\caption{Cumulative distributions of the translations errors and rotation errors on the two datasets. The x-axis represents the error and the y-axis denotes the percentage of image frames with error lower than the x-value.}
\label{fig:loopfull}
\end{figure*}

\textbf{RobotCar. } The RobotCar is especially challenging for the presence of weather variations, dynamic objects/pedestrians, occlusions, \etc. Following~\cite{kendall2015posenet, brahmbhatt2018geometry, xue2019local, xue2020learning}, we conduct experiments on the two subsets from the dataset. The LOOP route covers 1120m and the FULL route has a total length of 9562m.

As PoseNet~\cite{kendall2015posenet} conducts camera pose regression with heavy reliance on the visual information from singe images, large amounts of outliers are produced with insufficient inter-frame correlations, thus yielding low accuracy.
MapNet~\cite{brahmbhatt2018geometry} utilizes inputs from other sensors like GPS and IMU and fuses the measurements to aid the camera relocalization. Specifically, MapNet(pgo) acquires the relative camera pose from VO and acts in a sliding-window manner to predict the absolute poses.  
Compared with GNN-based approach~\cite{xue2020learning}, the proposed network shows major improvement as it efficiently models the spatiotemporal relation for sequential images, whereas the former network mainly relies on the spatial inter-frame dependencies.
\begin{wraptable}{r}{0.63\textwidth}
\midsepremove
\begin{center}
    \caption{Ablations on Pumpkin, Court and LOOP. }
    \label{table:ablation}
    \resizebox{\linewidth}{!}{%
    \begin{tabular}{lll|ccc}
    \toprule[0.5mm]
    \multicolumn{3}{l}{Config.} & Pumpkin & Court & LOOP\\
    \midrule[0.3mm]
    \multicolumn{3}{l}{TransCamP}& 0.12m, 1.82$^\circ$ & 4.57m, 2.03$^\circ$ & 3.27m, 1.99$^\circ$\\
    \multicolumn{3}{l}{TransCamp(temporal)}&0.33m, 3.11$^\circ$ & 8.77m, 5.3$^\circ$ & 4.22m, 2.58$^\circ$\\
    \multicolumn{3}{l}{TransCamp(graph)}&0.15m, 1.97$^\circ$ & 5.12m, 2.42$^\circ$ & 4.54m, 2.06$^\circ$\\
    \multicolumn{3}{l}{TransCamp(no MPNN)}&0.19m, 2.13$^\circ$ & 7.26m, 3.55$^\circ$ & 6.34m, 2.43$^\circ$\\
    \toprule[0.5mm]
    \end{tabular}
    }
\end{center}
\end{wraptable}
Additionally, we report the cumulative distributions of the translation and rotation prediction errors on the two datasets against prior work in Fig.~\ref{fig:loopfull}. The baselines include PoseNet~\cite{kendall2015posenet}, MapNet~\cite{brahmbhatt2018geometry}, LsG~\cite{xue2019local} and CNN+GNN~\cite{xue2020learning}. It can be observed that the proposed network outperforms the baselines on all the datasets.

\subsection{Ablation Study}
\label{sec:ablation}
We conduct ablation study to investigate the significance of different modules of the proposed network. We run the ablation experiments on `Pumpkin' scene from 7Scenes dataset, `Court' from the Cambridge and LOOP from the RobotCar dataset, to cover scenes of different scales and lengths. The results are given in Table.~\ref{table:ablation}.

We first evaluate TransCamP without the MPNN layers, such that the graph is directly fed into the graph transformer layers without the node information aggregation aided by the adjacency tensor. It can be observed that the performance of the network is significantly worse on the `Court' dataset. The reason is that the simple linear projection of the node features cannot preserve much information, compared with the message aggregated node features in the original network, where the neighboring node information is efficiently preserved. For the `Pumpkin' scene, high amounts of repetitive patterns are present such that the graph is densely connected; For the LOOP route, the images are highly consecutive such that temporal transformer can capture the neighboring node information along the temporal dimension.

We then study the effects of the individual transformer modules, \ie, the experiments are conducted with TransCamP without graph transformer layers (TransCamP(temporal)) and without temporal transformer layers (TransCamp(graph)). It can be observed that the accuracy of TransCamP(temporal) decreses harshly on `Court' and `Pumpkin' without the spatial correlation. Indeed, TransCamP can be seen as a GNN+RNN type of camera relocalization network, which can only preserve inter-frame dependencies over short period of time but tend to yield a overly sparse graph. On the other hand, the performance of TransCamP(graph) is slightly worse than the original network on all the datasets without significant decreased accuracy.   
\subsection{Discussions and Limitations}
\label{sec:limit}
From the experiments and the ablation study, we have observed that the output graphs are mostly very dense according to the spatiotemporal dependencies. The high density brings in high amounts of unnecessary computations, especially in the case where the scene scale is small and the camera motion is slow. Other challenging scenarios include when the camera motion is extremely fast, such that though the consecutive frames do not have sufficient overlapping views but it is difficult to remove the temporal edges, yielding large amounts of redundant edges. Equipping more GNN layers after the transformer layers can remove the unnecessary edges but tends to introduce over-fitting and graph memory overhead to the network. This work aims to develop a new framework for camera relocalization research and does not perceive potential negative societal impacts.

\section{Conclusion}
In this paper we propose a neural network approach with a graph transformer backbone, namely TransCamP, to address the camera relocalization problem.
We model the multi-view camera pose estimation problem with graph embedding, where the image features, camera poses and pair-wise camera transformations are fused into graph attributes.
With the introduction of a novel tensorized adjacency matrix, the proposed network can effectively capture the local node connection information.
By leveraging graph transformer layers with edge features and enabling temporal transformer to generate the spatiotemporal dependencies between the frames, TransCamP can actively gain the graph attention and achieves improved robustness, accuracy and efficiency.
\clearpage
{
\nocite{*}
\bibliographystyle{abbrvnat}
\bibliography{21nipsref}

\begin{thebibliography}{47}
\providecommand{\natexlab}[1]{#1}
\providecommand{\url}[1]{\texttt{#1}}
\expandafter\ifx\csname urlstyle\endcsname\relax
  \providecommand{\doi}[1]{doi: #1}\else
  \providecommand{\doi}{doi: \begingroup \urlstyle{rm}\Url}\fi

\bibitem[Balntas et~al.(2018)Balntas, Li, and Prisacariu]{balntas2018relocnet}
V.~Balntas, S.~Li, and V.~Prisacariu.
\newblock Relocnet: Continuous metric learning relocalisation using neural
  nets.
\newblock In \emph{European Conference on Computer Vision (ECCV)}, 2018.

\bibitem[Brahmbhatt et~al.(2018)Brahmbhatt, Gu, Kim, Hays, and
  Kautz]{brahmbhatt2018geometry}
S.~Brahmbhatt, J.~Gu, K.~Kim, J.~Hays, and J.~Kautz.
\newblock Geometry-aware learning of maps for camera localization.
\newblock In \emph{Proceedings of the IEEE Conference on Computer Vision and
  Pattern Recognition (CVPR)}, 2018.

\bibitem[Cai et~al.(2019)Cai, Ge, Liu, Cai, Cham, Yuan, and
  Thalmann]{cai2019exploiting}
Y.~Cai, L.~Ge, J.~Liu, J.~Cai, T.-J. Cham, J.~Yuan, and N.~M. Thalmann.
\newblock Exploiting spatial-temporal relationships for 3d pose estimation via
  graph convolutional networks.
\newblock In \emph{Proceedings of the IEEE International Conference on Computer
  Vision (ICCV)}, 2019.

\bibitem[Clark et~al.(2017)Clark, Wang, Markham, Trigoni, and
  Wen]{clark2017vidloc}
R.~Clark, S.~Wang, A.~Markham, N.~Trigoni, and H.~Wen.
\newblock Vid{L}oc: A deep spatio-temporal model for 6-dof video-clip
  relocalization.
\newblock In \emph{Proceedings of the IEEE Computer Society Conference on
  Computer Vision and Pattern Recognition (CVPR)}, 2017.

\bibitem[Crandall et~al.(2011)Crandall, Owens, Snavely, and
  Huttenlocher]{crandall2011discrete}
D.~Crandall, A.~Owens, N.~Snavely, and D.~Huttenlocher.
\newblock Discrete-continuous optimization for large-scale structure from
  motion.
\newblock In \emph{Proceedings of the IEEE Conference on Computer Vision and
  Pattern Recognition (CVPR)}. IEEE, 2011.

\bibitem[Dai et~al.(2017{\natexlab{a}})Dai, Chang, Savva, Halber, Funkhouser,
  and Nie{\ss}ner]{dai2017scannet}
A.~Dai, A.~X. Chang, M.~Savva, M.~Halber, T.~Funkhouser, and M.~Nie{\ss}ner.
\newblock Scan{N}et: Richly-annotated 3d reconstructions of indoor scenes.
\newblock In \emph{Proceedings of the IEEE Conference on Computer Vision and
  Pattern Recognition (CVPR)}, 2017{\natexlab{a}}.

\bibitem[Dai et~al.(2017{\natexlab{b}})Dai, Nie{\ss}ner, Zollh{\"o}fer, Izadi,
  and Theobalt]{dai2017bundlefusion}
A.~Dai, M.~Nie{\ss}ner, M.~Zollh{\"o}fer, S.~Izadi, and C.~Theobalt.
\newblock Bundlefusion: Real-time globally consistent 3d reconstruction using
  on-the-fly surface reintegration.
\newblock \emph{ACM Transactions on Graphics (ToG)}, 36\penalty0 (4):\penalty0
  1, 2017{\natexlab{b}}.

\bibitem[Deng et~al.(2009)Deng, Dong, Socher, Li, Li, and
  Fei-Fei]{deng2009imagenet}
J.~Deng, W.~Dong, R.~Socher, L.-J. Li, K.~Li, and L.~Fei-Fei.
\newblock Imagenet: A large-scale hierarchical image database.
\newblock In \emph{IEEE conference on computer vision and pattern recognition
  (CVPR)}. Ieee, 2009.

\bibitem[Devlin et~al.(2019)Devlin, Chang, Lee, and Toutanova]{devlin2018bert}
J.~Devlin, M.-W. Chang, K.~Lee, and K.~Toutanova.
\newblock Bert: Pre-training of deep bidirectional transformers for language
  understanding.
\newblock \emph{NAACL}, 2019.

\bibitem[Ding et~al.(2019)Ding, Wang, Sun, Shi, and Luo]{ding2019camnet}
M.~Ding, Z.~Wang, J.~Sun, J.~Shi, and P.~Luo.
\newblock Camnet: Coarse-to-fine retrieval for camera re-localization.
\newblock In \emph{Proceedings of the IEEE International Conference on Computer
  Vision (ICCV)}, 2019.

\bibitem[Dwivedi and Bresson(2021)]{dwivedi2020generalization}
V.~P. Dwivedi and X.~Bresson.
\newblock A generalization of transformer networks to graphs.
\newblock \emph{DLG-AAAI}, 2021.

\bibitem[Garg et~al.(2016)Garg, Bg, Carneiro, and Reid]{garg2016unsupervised}
R.~Garg, V.~K. Bg, G.~Carneiro, and I.~Reid.
\newblock Unsupervised {CNN} for single view depth estimation: Geometry to the
  rescue.
\newblock In \emph{European Conference on Computer Vision (ECCV)}. Springer,
  2016.

\bibitem[Geiger et~al.(2012)Geiger, Lenz, and Urtasun]{Geiger2012CVPR}
A.~Geiger, P.~Lenz, and R.~Urtasun.
\newblock Are we ready for autonomous driving? the {KITTI} vision benchmark
  suite.
\newblock In \emph{Proceedings of the IEEE Conference on Computer Vision and
  Pattern Recognition (CVPR)}, 2012.

\bibitem[Gidaris and Komodakis(2019)]{gidaris2019generating}
S.~Gidaris and N.~Komodakis.
\newblock Generating classification weights with {GNN} denoising autoencoders
  for few-shot learning.
\newblock In \emph{Proceedings of the IEEE Computer Society Conference on
  Computer Vision and Pattern Recognition (CVPR)}, 2019.

\bibitem[Gilmer et~al.(2017)Gilmer, Schoenholz, Riley, Vinyals, and
  Dahl]{gilmer2017neural}
J.~Gilmer, S.~S. Schoenholz, P.~F. Riley, O.~Vinyals, and G.~E. Dahl.
\newblock Neural message passing for quantum chemistry.
\newblock In \emph{International Conference on Machine Learning (ICML)}, 2017.

\bibitem[He et~al.(2016)He, Zhang, Ren, and Sun]{he2016deep}
K.~He, X.~Zhang, S.~Ren, and J.~Sun.
\newblock Deep residual learning for image recognition.
\newblock In \emph{Proceedings of the IEEE Conference on Computer Vision and
  Pattern Recognition (CVPR)}, 2016.

\bibitem[Kendall and Cipolla(2016)]{kendall2016modelling}
A.~Kendall and R.~Cipolla.
\newblock Modelling uncertainty in deep learning for camera relocalization.
\newblock In \emph{IEEE international conference on Robotics and Automation
  (ICRA)}. IEEE, 2016.

\bibitem[Kendall and Cipolla(2017)]{kendall2017geometric}
A.~Kendall and R.~Cipolla.
\newblock Geometric loss functions for camera pose regression with deep
  learning.
\newblock In \emph{Proceedings of the IEEE Conference on Computer Vision and
  Pattern Recognition (CVPR)}, 2017.

\bibitem[Kendall et~al.(2015)Kendall, Grimes, and Cipolla]{kendall2015posenet}
A.~Kendall, M.~Grimes, and R.~Cipolla.
\newblock Pose{N}et: A convolutional network for real-time 6-dof camera
  relocalization.
\newblock In \emph{Proceedings of the IEEE International Conference on Computer
  Vision (ICCV)}, 2015.

\bibitem[Kipf and Welling(2017)]{kipf2016semi}
T.~N. Kipf and M.~Welling.
\newblock Semi-supervised classification with graph convolutional networks.
\newblock \emph{International Conference on Learning Representations (ICLR)},
  2017.

\bibitem[Klodt and Vedaldi(2018)]{klodt2018supervising}
M.~Klodt and A.~Vedaldi.
\newblock Supervising the new with the old: learning sfm from sfm.
\newblock In \emph{Proceedings of the European Conference on Computer Vision
  (ECCV)}, 2018.

\bibitem[Laskar et~al.(2017)Laskar, Melekhov, Kalia, and
  Kannala]{laskar2017camera}
Z.~Laskar, I.~Melekhov, S.~Kalia, and J.~Kannala.
\newblock Camera relocalization by computing pairwise relative poses using
  convolutional neural network.
\newblock In \emph{Proceedings of the IEEE International Conference on Computer
  Vision Workshops}, 2017.

\bibitem[Maddern et~al.(2017{\natexlab{a}})Maddern, Pascoe, Linegar, and
  Newman]{maddern20171}
W.~Maddern, G.~Pascoe, C.~Linegar, and P.~Newman.
\newblock 1 year, 1000 km: The oxford robotcar dataset.
\newblock \emph{The International Journal of Robotics Research(IJRR)},
  36\penalty0 (1):\penalty0 3--15, 2017{\natexlab{a}}.

\bibitem[Maddern et~al.(2017{\natexlab{b}})Maddern, Pascoe, Linegar, and
  Newman]{maddern2017oxford}
W.~Maddern, G.~Pascoe, C.~Linegar, and P.~Newman.
\newblock {1 Year, 1000km: The Oxford RobotCar Dataset}.
\newblock \emph{The International Journal of Robotics Research (IJRR)},
  36\penalty0 (1):\penalty0 3--15, 2017{\natexlab{b}}.

\bibitem[Melekhov et~al.(2017)Melekhov, Ylioinas, Kannala, and
  Rahtu]{melekhov2017image}
I.~Melekhov, J.~Ylioinas, J.~Kannala, and E.~Rahtu.
\newblock Image-based localization using hourglass networks.
\newblock In \emph{Proceedings of the IEEE international conference on computer
  vision workshops (ICCV Workshops)}, 2017.

\bibitem[Purkait et~al.(2020)Purkait, Chin, and Reid]{purkait2020neurora}
P.~Purkait, T.-J. Chin, and I.~Reid.
\newblock Neurora: Neural robust rotation averaging.
\newblock In \emph{European Conference on Computer Vision (ECCV)}. Springer,
  2020.

\bibitem[Rong et~al.(2020)Rong, Huang, Xu, and Huang]{rong2019dropedge}
Y.~Rong, W.~Huang, T.~Xu, and J.~Huang.
\newblock Dropedge: Towards deep graph convolutional networks on node
  classification.
\newblock \emph{International Conference on Learning Representations (ICLR)},
  2020.

\bibitem[Sattler et~al.(2019)Sattler, Zhou, Pollefeys, and
  Leal-Taixe]{sattler2019understanding}
T.~Sattler, Q.~Zhou, M.~Pollefeys, and L.~Leal-Taixe.
\newblock Understanding the limitations of cnn-based absolute camera pose
  regression.
\newblock In \emph{Proceedings of the IEEE Computer Society Conference on
  Computer Vision and Pattern Recognition (CVPR)}, 2019.

\bibitem[Scarselli et~al.(2008)Scarselli, Gori, Tsoi, Hagenbuchner, and
  Monfardini]{scarselli2008graph}
F.~Scarselli, M.~Gori, A.~C. Tsoi, M.~Hagenbuchner, and G.~Monfardini.
\newblock The graph neural network model.
\newblock \emph{IEEE Transactions on Neural Networks}, 20\penalty0
  (1):\penalty0 61--80, 2008.

\bibitem[Shotton et~al.(2013)Shotton, Glocker, Zach, Izadi, Criminisi, and
  Fitzgibbon]{shotton2013scene}
J.~Shotton, B.~Glocker, C.~Zach, S.~Izadi, A.~Criminisi, and A.~Fitzgibbon.
\newblock Scene coordinate regression forests for camera relocalization in
  rgb-d images.
\newblock In \emph{Proceedings of the IEEE Conference on Computer Vision and
  Pattern Recognition (CVPR)}, 2013.

\bibitem[Tang and Tan(2018)]{tang2018ba}
C.~Tang and P.~Tan.
\newblock {BA-N}et: Dense bundle adjustment networks.
\newblock In \emph{International Conference on Learning Representations
  (ICLR)}, 2018.

\bibitem[Triggs et~al.(1999)Triggs, McLauchlan, Hartley, and
  Fitzgibbon]{triggs1999bundle}
B.~Triggs, P.~F. McLauchlan, R.~I. Hartley, and A.~W. Fitzgibbon.
\newblock Bundle adjustment—a modern synthesis.
\newblock In \emph{International Workshop on Vision Algorithms}. Springer,
  1999.

\bibitem[Valada et~al.(2018)Valada, Radwan, and Burgard]{valada2018deep}
A.~Valada, N.~Radwan, and W.~Burgard.
\newblock Deep auxiliary learning for visual localization and odometry.
\newblock In \emph{Proceedings of the IEEE International Conference on Robotics
  and Automation (ICRA)}, 2018.

\bibitem[Vaswani et~al.(2017)Vaswani, Shazeer, Parmar, Uszkoreit, Jones, Gomez,
  Kaiser, and Polosukhin]{vaswani2017attention}
A.~Vaswani, N.~Shazeer, N.~Parmar, J.~Uszkoreit, L.~Jones, A.~N. Gomez,
  L.~Kaiser, and I.~Polosukhin.
\newblock Attention is all you need.
\newblock \emph{Conference on Neural Information Processing Systems (NeurIPS)},
  2017.

\bibitem[Veli{\v{c}}kovi{\'c} et~al.(2018)Veli{\v{c}}kovi{\'c}, Cucurull,
  Casanova, Romero, Lio, and Bengio]{velivckovic2017graph}
P.~Veli{\v{c}}kovi{\'c}, G.~Cucurull, A.~Casanova, A.~Romero, P.~Lio, and
  Y.~Bengio.
\newblock Graph attention networks.
\newblock \emph{International Conference on Learning Representations (ICLR)},
  2018.

\bibitem[Vijayanarasimhan et~al.(2017)Vijayanarasimhan, Ricco, Schmid,
  Sukthankar, and Fragkiadaki]{vijayanarasimhan2017sfm}
S.~Vijayanarasimhan, S.~Ricco, C.~Schmid, R.~Sukthankar, and K.~Fragkiadaki.
\newblock Sfm-net: Learning of structure and motion from video.
\newblock \emph{arXiv preprint arXiv:1704.07804}, 2017.

\bibitem[Walch et~al.(2017)Walch, Hazirbas, Leal-Taixe, Sattler, Hilsenbeck,
  and Cremers]{walch2017image}
F.~Walch, C.~Hazirbas, L.~Leal-Taixe, T.~Sattler, S.~Hilsenbeck, and
  D.~Cremers.
\newblock Image-based localization using lstms for structured feature
  correlation.
\newblock In \emph{Proceedings of the IEEE International Conference on Computer
  Vision (ICCV)}, 2017.

\bibitem[Wang et~al.(2019)Wang, Ji, Shi, Wang, Ye, Cui, and
  Yu]{wang2019heterogeneous}
X.~Wang, H.~Ji, C.~Shi, B.~Wang, Y.~Ye, P.~Cui, and P.~S. Yu.
\newblock Heterogeneous graph attention network.
\newblock In \emph{The World Wide Web Conference (WWW)}, 2019.

\bibitem[Wu et~al.(2011)]{wu2011visualsfm}
C.~Wu et~al.
\newblock Visualsfm: A visual structure from motion system.
\newblock 2011.

\bibitem[Wu et~al.(2017)Wu, Ma, and Hu]{wu2017delving}
J.~Wu, L.~Ma, and X.~Hu.
\newblock Delving deeper into convolutional neural networks for camera
  relocalization.
\newblock In \emph{IEEE International Conference on Robotics and Automation
  (ICRA)}. IEEE, 2017.

\bibitem[Xu et~al.(2019)Xu, Hu, Leskovec, and Jegelka]{xu2018powerful}
K.~Xu, W.~Hu, J.~Leskovec, and S.~Jegelka.
\newblock How powerful are graph neural networks?
\newblock \emph{International Conference on Learning Representations (ICLR)},
  2019.

\bibitem[Xue et~al.(2019)Xue, Wang, Yan, Wang, Wang, and Zha]{xue2019local}
F.~Xue, X.~Wang, Z.~Yan, Q.~Wang, J.~Wang, and H.~Zha.
\newblock Local supports global: Deep camera relocalization with sequence
  enhancement.
\newblock In \emph{Proceedings of the IEEE International Conference on Computer
  Vision (ICCV)}, 2019.

\bibitem[Xue et~al.(2020)Xue, Wu, Cai, and Wang]{xue2020learning}
F.~Xue, X.~Wu, S.~Cai, and J.~Wang.
\newblock Learning multi-view camera relocalization with graph neural networks.
\newblock In \emph{Proceedings of the IEEE Computer Society Conference on
  Computer Vision and Pattern Recognition (CVPR)}, 2020.

\bibitem[Yun et~al.(2019)Yun, Jeong, Kim, Kang, and Kim]{yun2019graph}
S.~Yun, M.~Jeong, R.~Kim, J.~Kang, and H.~J. Kim.
\newblock Graph transformer networks.
\newblock \emph{Conference on Neural Information Processing Systems (NeurIPS)},
  2019.

\bibitem[Zhang et~al.(2020)Zhang, Zhang, Xia, and Sun]{zhang2020graph}
J.~Zhang, H.~Zhang, C.~Xia, and L.~Sun.
\newblock Graph-bert: Only attention is needed for learning graph
  representations.
\newblock \emph{arXiv preprint arXiv:2001.05140}, 2020.

\bibitem[Zhang and Chen(2018)]{zhang2018link}
M.~Zhang and Y.~Chen.
\newblock Link prediction based on graph neural networks.
\newblock \emph{Conference on Neural Information Processing Systems (NeurIPS)},
  2018.

\bibitem[Zhao and Akoglu(2019)]{zhao2019pairnorm}
L.~Zhao and L.~Akoglu.
\newblock Pair{N}orm: Tackling oversmoothing in gnns.
\newblock In \emph{International Conference on Learning Representations
  (ICLR)}, 2019.

\end{thebibliography}
}
\end{document}